\documentclass[twoside]{article}
\pdfoutput=1

\usepackage[preprint]{aistats2026}
\usepackage{amsmath,amssymb,amsthm,mathtools}
\usepackage{booktabs}
\usepackage{graphicx}
\usepackage{hyperref}
\hypersetup{hidelinks}
\usepackage{microtype}
\usepackage{placeins}
\usepackage[round]{natbib}

\newtheorem{theorem}{Theorem}
\newtheorem{proposition}{Proposition}
\newtheorem{lemma}{Lemma}
\newtheorem{remark}{Remark}

\newcommand{\R}{\mathbb{R}}
\newcommand{\C}{\mathbb{C}}
\newcommand{\E}{\mathbb{E}}
\newcommand{\norm}[1]{\left\lVert #1 \right\rVert}
\newcommand{\spec}{\operatorname{spec}}

\begin{document}

\runningtitle{Eigenbasis-Independent Learnable Spectral PEs for Directed Graphs}

\twocolumn[

\aistatstitle{Eigenbasis-Independent Learnable Spectral Positional Encodings for
Directed Graphs via Hermitian Block Krylov Subspaces}

\aistatsauthor{Jiaqing Xie \And Yuxin Wang}

\aistatsaddress{Fudan University \And Fudan University} ]

\begin{abstract}
Spectral positional encodings (PEs) for \emph{directed} graphs face two
obstacles: full-spectrum magnetic methods require a dense Hermitian
eigendecomposition per potential, and complex eigenvectors are defined
only up to basis choices within eigenspaces, which prior work handles with basis-invariant
architectures. We propose learnable spectral PEs of the form
$h_\theta(A_q)\,R$, where $A_q$ is a normalized magnetic operator,
$h_\theta$ a learnable scalar spectral response, and $R$ a block of random
probes. Because the PE is a \emph{matrix function} of the operator, it is
independent of eigendecomposition conventions. We compute it in a Hermitian block Krylov
subspace from sparse matrix--vector products only, prove that
$k = O(\log(1/\varepsilon))$ block steps suffice uniformly over
heat--resolvent response families, and give a covering-number argument for
why low-dimensional structured families generalize where an unconstrained
per-eigenvalue oracle overfits. On a directed SBM whose symmetrization is
uninformative by construction, direction-blind PEs stay at chance while
magnetic Krylov PEs converge to the exact-eigendecomposition oracle as the
depth grows. Cross-response probe inner products provide pairwise features
with $1/\sqrt{s}$ Monte-Carlo error, and the undirected $q{=}0$ case improves
heterophilous benchmarks over no-PE and polynomial baselines.
\end{abstract}

\section{Introduction}

Positional encodings are a key ingredient of graph transformers and a
practical route to lifting the expressive power of message-passing GNNs
beyond 1-WL \citep{dwivedi2020generalization,rampavsek2022recipe}. The dominant
spectral construction, Laplacian PE, uses eigenvectors of the symmetric
graph Laplacian and inherits three well-known problems: the cost of extracting
and stabilizing individual eigenvectors, instability under spectral perturbation, and sign/basis
ambiguity of eigenvectors \citep{lim2023sign,wang2022equivariant,
huang2024stability}. Learnable spectral PEs (LLPE) replace raw eigenvectors
with a trainable filter of the spectrum \citep{pmlr-v258-ito25a}, which improves
task adaptivity but still requires a spectral representation on which the
filter is applied.

Directed graphs sharpen every one of these issues. Direction matters in
citation, program, and circuit graphs, yet the adjacency matrix is no longer
symmetric, and the standard fix of symmetrization destroys the
information of interest. The magnetic Laplacian \citep{geisler2023transformers}
restores symmetry in the form of a complex Hermitian operator $L_q$ whose
phases encode edge directions, and Multi-$q$ magnetic PEs
\citep{huang2025good} show that a \emph{set} of potentials $q$ is provably
necessary to express directed walk profiles. Practical eigenvector-based
magnetic PEs use a partial Hermitian eigensolver \emph{per potential}, while
exact full-spectrum pairwise readouts require a dense decomposition; in both
cases, complex eigenvectors are defined only up to per-eigenspace unitary gauge,
which \citet{huang2025good} address with dedicated basis-invariant
architectures.

\paragraph{This paper.} We take a different route. We define the PE directly
as a learnable \emph{matrix function} of the magnetic operator applied to a
block of random probes,
\begin{equation}
\label{eq:pe}
Z_q(\theta) \;=\; h_\theta(A_q)\,R \in \C^{n \times s},
\qquad R \sim \mathcal{CN}(0, \tfrac{1}{s} I),
\end{equation}
and approximate it in a Hermitian block Krylov subspace
$\mathcal{K}_k(A_q, R)$ using sparse matrix--vector products only. This
design has three consequences. The probes are part of the randomized
encoding: for a fixed draw, equivariance is conditional on permuting $R$ with
the nodes; if probes are freshly sampled after relabeling, the encoding is
permutation equivariant in distribution because the Gaussian law is
exchangeable (Proposition~\ref{prop:probe}).
Here ``gauge invariance'' refers only to eigenbasis choices. Under a
vertex-wise magnetic gauge, the PE is covariant if $R$ co-transforms.

\textbf{(1) Independence from eigenbasis choices.} A matrix
function $h(A_q)$ is defined independently of any eigendecomposition;
eigenvectors, their phases, and their basis choices never appear. The
invariance that prior work recovers with special networks
\citep{lim2023sign,huang2025good} holds here by construction
(Proposition~\ref{prop:gauge}).

\textbf{(2) Approximation is provably cheap, uniformly over the learnable
family.} For structured response families built from heat kernels and
resolvents, we show the block Krylov approximation error decays exponentially
in the number of steps $k$, \emph{uniformly} over the family parameters
(Theorem~\ref{thm:approx}). The learnable PE family computed in $k$ Krylov
steps exactly realizes degree-$(k-1)$ scalar Chebyshev responses on the probes,
the same response class used by LLPE, and
$\varepsilon$-covers the analytic family with
$k = O(\log(1/\varepsilon))$.

\textbf{(3) Structure controls estimation.} LLPE uses a shared Chebyshev
mapping whose parameter count is independent of graph size
\citep{pmlr-v258-ito25a}. We likewise study low-dimensional structured
responses, and contrast them with an \emph{unconstrained diagnostic oracle}
that assigns one weight to each eigenvalue. The latter creates an
$n$-parameter hypothesis class per potential. A covering-number argument
(Proposition~\ref{prop:gen}) and matched experiments show why this oracle can
overfit in the low-label regime.

Empirically, we construct a cyclic directed SBM whose symmetrized graph is
(approximately) a homogeneous Erd\H{o}s--R\'enyi graph, so \emph{all} class
signal lives in edge directions. Direction-blind PEs ($q{=}0$) are at chance;
magnetic Krylov PEs recover the structure and converge monotonically to the
exact-eigendecomposition oracle as $k$ grows. On real directed
node-classification benchmarks (WebKB, Wikipedia networks) we find that no
spectral PE beats random probes. We report this negative result and take it
as evidence that these benchmarks may not strongly reward the type of global
directional information captured by magnetic spectral PEs under our setup,
which further motivates controlled diagnostics.

\section{Related Work}

\paragraph{Spectral PEs and their invariance problem.} Laplacian PEs
\citep{dwivedi2020generalization} require eigendecomposition and suffer
sign/basis ambiguity, addressed architecturally by SignNet/BasisNet
\citep{lim2023sign} and stable variants \citep{huang2024stability,
wang2022equivariant}. PEARL \citep{kanatsoulis2025learning} approximates
equivariant functions of eigenvectors with message-passing and random or
basis inputs, achieving linear complexity. Random feature propagation
\citep{eliasof2023graph} concatenates iterates of random features under
predefined or learned graph-dependent propagation operators. Its fixed-operator
form spans the same power sequence that underlies a Krylov space. We therefore
include a matched magnetic RFP baseline; our block Krylov cache orthogonalizes
and compresses this span and supports repeated evaluation of non-polynomial
response families.

\paragraph{Random spectral kernels.} Prior work GIST
\citep{rigotti2026gist} studies embeddings $f(P)R$ for real symmetric graph
operators and uses their inner products as unbiased estimators of
$f(P)f(P)^\top$, with $O(s^{-1/2})$ statistical error, in scalable spectral
attention and neural operators. We use this random spectral-kernel principle
rather than claim it as new. Our focus is the learnable directed setting:
complex Hermitian magnetic operators, multiple potentials, cross-response
pairwise readouts, and a reusable block-Krylov cache with approximation
guarantees uniform over the response parameters.

\paragraph{Learnable spectral filters.} Spectral GNNs learn polynomial
\citep{defferrard2016convolutional,he2021bernnet,wang2022powerful} or
set-to-set \citep{bo2023specformer} filters; LLPE \citep{pmlr-v258-ito25a}
uses a shared truncated Chebyshev mapping over the full spectrum.
LanczosNet \citep{liao2018lanczosnet} used a
Lanczos basis inside a GNN, without random probes, directed operators, or
approximation guarantees for a learnable family.

\paragraph{Directed graph PEs.} MagNet \citep{zhang2021magnet} and magnetic
PEs \citep{geisler2023transformers} introduced magnetic Laplacians to graph
learning; Multi-$q$ Mag-PE \citep{huang2025good} showed multiple potentials
are needed to express walk profiles and built basis-invariant networks over
complex eigenvectors. Practical versions retain the lowest $k_{\rm eig}$
eigenpairs using a partial eigensolver; full-spectrum versions become dense
oracles. Our PE inherits the multi-$q$ insight but replaces
eigenvectors with eigenbasis-independent matrix functions computed by sparse
matvecs.

\paragraph{Krylov methods for matrix functions.} The convergence of (block)
Lanczos approximations to $f(A)b$ is classical
\citep{saad1992analysis,musco2018stability}: the error is governed by the
best uniform polynomial approximation of $f$ on the spectral interval. Our
Theorem~\ref{thm:approx} is an application of this machinery made
uniform over a learnable response family, as required for a learning
guarantee.

\begin{table*}[t]
\centering
\caption{Comparison with closely related spectral and random-feature PEs.
``Sparse only'' means no dense eigendecomposition of the original graph
operator. ``Basis independent'' means independence from eigenvector choices,
not invariance under vertex-wise magnetic gauge transformations.}
\label{tab:related_comparison}
\resizebox{\textwidth}{!}{%
\begin{tabular}{lcccccc}
\toprule
Method & Directed & Learnable & Eigenvector-free & Basis independent & Sparse only & Pairwise entries \\
\midrule
RFP \citep{eliasof2023graph}
  & operator-dependent & operator & yes & yes & yes & no \\
GIST \citep{rigotti2026gist}
  & no & filter/head & yes & via inner products & yes & yes \\
PEARL \citep{kanatsoulis2025learning}
  & no & no & yes & via network & yes & no \\
LanczosNet \citep{liao2018lanczosnet}
  & no & yes & partial & no & partial & no \\
LLPE \citep{pmlr-v258-ito25a}
  & no & Chebyshev & no & sign handling & no & no \\
Multi-$q$ Mag-PE \citep{huang2025good}
  & yes & via head & no & via SPE & no & yes \\
Mag-Krylov (ours)
  & yes & response & yes & matrix function & yes & yes \\
\bottomrule
\end{tabular}
}
\end{table*}

\section{Method}
\label{sec:method}

\paragraph{Setup.} Let $G = (V, E)$ be a directed graph with $n = |V|$,
adjacency $a_{uv} \in \{0,1\}$ and no self-loops. For a potential
$q \in [0, \tfrac12]$ define the symmetrized weights and phases
\begin{align*}
a^{\mathrm{sym}}_{uv} &= \tfrac{1}{2}(a_{uv} + a_{vu}), &
\Theta^{(q)}_{uv} &= 2\pi q\,(a_{uv} - a_{vu}),
\end{align*}
the Hermitian matrix $H_q(u,v) = a^{\mathrm{sym}}_{uv}
e^{i\Theta^{(q)}_{uv}}$, the degree $d_u = \sum_v a^{\mathrm{sym}}_{uv}$, and
the normalized operator
\begin{equation}
A_q \;=\; -\,D^{-1/2} H_q D^{-1/2}, \qquad \spec(A_q) \subseteq [-1, 1].
\end{equation}
$A_q = L_q - I$ where $L_q$ is the normalized magnetic Laplacian;
$q = 0$ recovers the symmetrized (direction-blind) operator.

\paragraph{Learnable spectral responses.} A response family is a set
$\mathcal{H} = \{h_\theta : [-1,1] \to \R,\ \theta \in \Theta\}$. We study:
\emph{(a) Chebyshev}: $h_\theta(\xi) = \sum_{m=0}^{M} c_m T_m(\xi)$;
\emph{(b) heat--resolvent mixtures} on $\mu = \xi + 1 \in [0,2]$:
\begin{equation}
\label{eq:hr}
h_\theta(\xi) = \beta + \sum_{j=1}^{m} \alpha_j\, e^{-t_j \mu}
              + \sum_{j=1}^{m} \gamma_j\, (\mu + \tau_j)^{-1},
\end{equation}
with $t_j \in [t_{\min}, t_{\max}]$, $\tau_j \ge \tau_{\min} > 0$;
\emph{(c) MLP}: $h_\theta(\xi) = \mathrm{MLP}(\phi(\xi))$ with fixed Fourier
features $\phi$; and \emph{(d) an unconstrained spectral oracle}: one
independent parameter per eigenvalue, used only as a capacity diagnostic.
The Chebyshev family is the direct analogue of the shared response used by
LLPE \citep{pmlr-v258-ito25a}. Per-head RMS normalization of $h_\theta$ over
the spectrum, with
a learnable gain, puts all families on a common scale; we found this
essential for fair comparisons (Section~\ref{sec:exp}).

\paragraph{Hermitian block Krylov approximation.} Given complex Gaussian
probes $R \in \C^{n \times s}$, $k$ steps of QR-stabilized block Lanczos with
full reorthogonalization produce an orthonormal basis
$Q \in \C^{n \times r}$ ($r \le ks$) of
$\mathcal{K}_k(A_q, R) = \mathrm{span}\{R, A_q R, \dots, A_q^{k-1} R\}$,
the projected operator $T = Q^{\mathsf H} A_q Q$, and $G = Q^{\mathsf H} R$.
The PE for potential $q$ is
\begin{equation}
\label{eq:krylovpe}
\widehat{Z}_q(\theta) = Q\, h_\theta(T)\, G
= Q V h_\theta(\Xi) V^{\mathsf H} G,
\end{equation}
where $T = V \Xi V^{\mathsf H}$ is the $r \times r$ Hermitian
eigendecomposition, computed once in $O(r^3) = O(k^3 s^3)$, independent of
$n$. Multiple potentials $\vec q = (q_1, \dots, q_Q)$ are handled with one
Krylov cache each; features $[\mathrm{Re}\,\widehat Z_{q_i};\,
\mathrm{Im}\,\widehat Z_{q_i}]$ are concatenated over potentials and heads
and projected to the PE dimension. Training only differentiates through
$H$ response heads on the fixed cache, so each epoch costs
$O(QH\,(r^2 s + n r s))$.
The total precompute is $O(Q\,k\,(\mathrm{nnz}(A) \,s + n k s^2))$ sparse
work, and no dense $n \times n$ object is ever formed. During response
training, retaining $Q,T,G$ for every potential requires
$O(Qnr+Qr^2+Qrs)=O(Qnks+Qk^2s^2)$ storage in the generic no-deflation case.
After freezing the responses, one may instead materialize only the final
$O(QHns)$ filtered features, or their $O(nd_{\rm PE})$ learned projection.

\section{Theory}
\label{sec:theory}

Throughout, $A \in \C^{n\times n}$ is Hermitian with $\spec(A) \subseteq
[-1,1]$, $R \in \C^{n \times s}$, and $Q, T, G$ are produced by $k$ exact
block Lanczos steps. Proofs are in Appendix~\ref{app:proofs}.

\begin{proposition}[Eigenbasis independence and equivariance]
\label{prop:gauge}
$Z_q(\theta) = h_\theta(A_q) R$ depends on $A_q$ only through the operator
itself: it is invariant to any choice of eigenbasis (including per-eigenspace
unitary gauge transformations of degenerate eigenspaces and global phase).
Moreover, for any permutation matrix $P$,
$h_\theta(P A_q P^\top) (PR) = P\, Z_q(\theta)$, and the same statements hold
for the Krylov approximation \eqref{eq:krylovpe}.
For a vertex-wise unitary $U$, replacing $(A_q,R)$ by
$(UA_qU^{\mathsf H},UR)$ maps the output to $UZ_q(\theta)$; this is covariance,
not invariance, under magnetic gauge transformations.
\end{proposition}

Unlike eigenvector PEs, there is no sign, phase, or basis ambiguity to
repair, because no eigenvector is ever selected. The
random probes should not be confused with deterministic node features:
Proposition~\ref{prop:probe} states the two precise equivariance notions we
use in experiments.

\begin{proposition}[Random probes and pairwise entries]
\label{prop:probe}
Let $R = [r_1,\ldots,r_s]$ have iid complex Gaussian columns with
$\E r_t r_t^{\mathsf H}=\sigma^2 I$, and let $F_a=h_a(A)$ and
$F_b=h_b(A)$ be bounded matrix functions of the same Hermitian operator.
\begin{enumerate}
\item For a fixed probe draw, $F_a(PAP^\top)(PR)=P F_a(A)R$. If instead
$R'$ is freshly sampled after relabeling, $F_a(PAP^\top)R' \stackrel{d}{=}
P F_a(A)R$.
\item For any node pair $(i,j)$,
\begin{equation}
\widehat K_{ab}(i,j)=\frac{1}{s\sigma^2} \sum_{t=1}^{s}
(F_a r_t)_i\,\overline{(F_b r_t)_j}
\end{equation}
is an unbiased estimator of $(F_aF_b^{\mathsf H})_{ij}$ with variance
$O(\norm{e_i^{\mathsf T}F_a}_2^2\norm{e_j^{\mathsf T}F_b}_2^2/s)$.
\end{enumerate}
\end{proposition}

The pairwise estimator is the complex cross-response analogue of the random
spectral-kernel estimator used by GIST \citep{rigotti2026gist}. Our use of it
is specific to multiple magnetic potentials and pairs of learned responses
$h_a,h_b$.

Raw probe coordinates do not themselves remove the probe-basis ambiguity:
for any right-unitary $U_s$, replacing $R$ by $R U_s$ maps
$h(A)R$ to $h(A)R U_s$. A coordinate-wise MLP is not generally invariant to
this action. We therefore use fixed probe coordinates for the transductive
node-classification experiments and make no cross-graph transfer claim for
them. The Gram and cross-response readouts in
Proposition~\ref{prop:probe} eliminate this ambiguity and are the appropriate
interface for inductive pairwise prediction.

\begin{lemma}[Block polynomial exactness]
\label{lem:exact}
For every polynomial $p$ of degree at most $k-1$,
$\;Q\,p(T)\,Q^{\mathsf H} R = p(A)\,R$.
\end{lemma}

\begin{proposition}[Directed walk-profile recovery]
\label{prop:walk}
Define the positive-sign normalized magnetic adjacency
$B_q=-A_q=D^{-1/2}H_qD^{-1/2}$. Let $h_a(x)=x^a$ and $h_b(x)=x^b$ with
$0\le a,b\le k-1$. For each potential $q$,
the block-Krylov cross-response estimator is an unbiased estimator of
$(A_q^{a+b})_{ij}=(-1)^{a+b}(B_q^{a+b})_{ij}$, and its standard deviation is
$O(s^{-1/2})$.
Consequently, for every $m\le 2k-2$, evaluating a sufficient Fourier grid of
potentials and multiplying by the known global factor $(-1)^m$ recovers the
degree-normalized length-$m$ directed walk profile between $(i,j)$ in
expectation, with entrywise finite-probe error $O(s^{-1/2})$.
\end{proposition}

The last step uses the Fourier correspondence between powers of magnetic
graph matrices and directed walk profiles \citep{huang2026powers}. For order
$m$, exact inversion requires $\lfloor m/2\rfloor+1$ distinct frequencies
under the convention of \citet{huang2026powers}; fewer potentials give a
compressed approximation. Unlike a generic analytic response, the two
polynomial actions incur zero Krylov approximation error by
Lemma~\ref{lem:exact}.

\begin{theorem}[Uniform Krylov approximation]
\label{thm:approx}
Let $\mathcal{H}$ be any response family and let
$E_{k-1}(h) = \inf_{\deg p \le k-1} \sup_{x \in [-1,1]} |h(x) - p(x)|$.
Then
\begin{equation}
\sup_{\theta \in \Theta}
\norm{h_\theta(A) R - Q h_\theta(T) G}_F
\;\le\; 2 \norm{R}_F\, \sup_{\theta \in \Theta} E_{k-1}(h_\theta).
\end{equation}
For the heat--resolvent family \eqref{eq:hr} with
$t_j \le t_{\max}$, $\tau_j \ge \tau_{\min}$, and mixture coefficients
bounded by $B$ in $\ell_1$,
\begin{equation}
\sup_{\theta} E_{k-1}(h_\theta)
\;\le\; C(B, t_{\max}, \tau_{\min})\; \rho^{\,k},
\qquad \rho < 1,
\end{equation}
where $\rho = \big(1 + \tau_{\min} + \sqrt{\tau_{\min}^2 + 2\tau_{\min}}
\big)^{-1}$ for the resolvent part and the heat part decays
super-geometrically. Consequently $k = O\!\big(\log(1/\varepsilon)\big)$
block steps suffice for a uniform $\varepsilon \norm{R}_F$ approximation of
the entire learnable PE family.
\end{theorem}

\begin{remark}
Chebyshev responses of degree $M \le k-1$ are reproduced \emph{exactly}
(Lemma~\ref{lem:exact}). Thus, when applied to random probes, the Krylov
construction exactly realizes the same degree-$(k-1)$ scalar Chebyshev
response class used by LLPE, while also $\varepsilon$-covering the analytic
families. This does not identify the resulting coordinates with LLPE's
eigenvector representation. In floating point, full reorthogonalization keeps the bounds
meaningful \citep{musco2018stability}; our implementation tracks
orthogonality and projection residuals explicitly.
\end{remark}

\begin{proposition}[Capacity of structured responses]
\label{prop:gen}
Fix compact response-parameter sets and a downstream predictor class whose
loss is bounded, uniformly Lipschitz in the PE, and has covering entropy
$\mathcal C_{\rm down}(\varepsilon)$ independent of the response
parameterization. For heat--resolvent responses with $m$ components per
potential, the joint class has covering entropy
$\mathcal C_{\rm down}(\varepsilon)+O(Qm\log(C/\varepsilon))$.
Consequently, up to the common downstream term, its empirical-process
deviation on $\ell$ labeled nodes scales as
$\widetilde O(\sqrt{Qm/\ell})$. An unconstrained per-eigenvalue response over
rank $r$ has $Qr$ free response parameters. Applying the same covering
argument gives an additional upper entropy term
$O(Qr\log(C/\varepsilon))$ and hence the worst-case incremental deviation bound
$\widetilde O(\sqrt{Qr/\ell})$, with $r$ as large as $n$. This comparison
concerns the unconstrained diagnostic oracle, not LLPE's shared Chebyshev
parameterization.
\end{proposition}

Proposition~\ref{prop:gen} separates the two error sources: the Krylov depth
$k$ controls the \emph{approximation bias} (Theorem~\ref{thm:approx}) while
the response family controls the \emph{estimation variance}, and the two can
be tuned independently.

\section{Experiments}
\label{sec:exp}

\paragraph{Naming convention.} Variants are named
\emph{operator-solver-family}. Operators: \textbf{Mag} (magnetic, grid
$\vec q$) and \textbf{Sym} ($q{=}0$, direction-blind); the prefix is dropped
in the undirected experiments of Section~\ref{sec:undirected}, where the
operator is the real symmetric $A = L - I$. Solvers: \textbf{Direct}
(explicit matrix recursion on $A$), \textbf{Krylov}
(Eq.~\ref{eq:krylovpe}), and \textbf{Exact} (dense eigendecomposition
oracle). Families: \textbf{Cheb}, \textbf{Heat}, \textbf{HR}
(heat--resolvent), \textbf{MLP}, and \textbf{Free} (the unconstrained
per-eigenvalue diagnostic oracle); see Section~\ref{sec:method}. For example, Mag-Krylov-HR
is the heat--resolvent family on magnetic operators computed in the Krylov
subspace.

\paragraph{Matched RFP baseline.} To isolate the effect of the block-Krylov
cache and spectral readout from random feature propagation, Magnetic-RFP
concatenates $[R,A_qR,\ldots,A_q^{k-1}R]$ over the same potentials and applies
a learned projection to the same PE dimension. It uses the same probes,
$k$, and downstream MLP, but no orthogonalization or small-matrix spectral
response. Its projection gives it more trainable parameters than
Mag-Krylov-HR in the main setting (63.8k vs.\ 27.1k), making it a conservative
capacity comparison.

\paragraph{A controlled directed benchmark.} We build a cyclic directed SBM
(\textsc{dsbm}): $C{=}3$ classes on a directed cycle; forward edges
($c \to c{+}1$) appear with probability $p_f$, backward with $p_b \ll p_f$,
and intra-class with $(p_f + p_b)/2$. By construction every unordered pair is
connected in the symmetrized graph with probability $\approx p_f + p_b$
regardless of classes: \emph{the symmetrized graph carries no class signal}.
Nodes have no features; the classifier sees the PE only. We use $n = 600$,
$s = 32$ probes, potentials $\vec q = (0, \tfrac16, \tfrac13)$, $k = 10$
steps, 5 seeds, and report test accuracy at best validation.

\begin{table}[t]
\centering
\caption{Directed SBM ($n{=}600$, 3 classes, chance ${\approx}33.3\%$).
Direction-blind PEs are at chance by construction; the matched Magnetic-RFP
baseline isolates raw directed propagation, while learned magnetic Krylov PEs
recover the planted structure. Oracles use dense eigendecompositions.}
\label{tab:main}
\resizebox{\columnwidth}{!}{\begin{tabular}{lcc}
\toprule
PE variant & Test accuracy & Precompute (s) \\
\midrule
Random probes & 31.9 $\pm$ 1.1 & 0.00 \\
LapPE (sym.\ eigvecs) & 32.6 $\pm$ 1.5 & 2.04 \\
RWSE (sym.\ walks) & 33.4 $\pm$ 0.2 & 0.21 \\
Mag-PE (eigvecs) & 51.1 $\pm$ 2.3 & 6.50 \\
\midrule
Sym-RFP ($q{=}0$) & 32.4 $\pm$ 1.6 & 0.02 \\
Sym-Krylov-Cheb ($q{=}0$) & 33.0 $\pm$ 2.0 & 0.04 \\
Sym-Krylov-HR ($q{=}0$) & 34.5 $\pm$ 2.5 & 0.04 \\
Sym-Krylov-MLP ($q{=}0$) & 31.1 $\pm$ 2.2 & 0.04 \\
\midrule
Magnetic-RFP & 38.4 $\pm$ 3.2 & 0.07 \\
Mag-Krylov-Cheb & 83.2 $\pm$ 7.8 & 0.12 \\
Mag-Krylov-Heat & \textbf{98.8 $\pm$ 0.7} & 0.12 \\
Mag-Krylov-HR & 97.6 $\pm$ 0.8 & 0.12 \\
Mag-Krylov-MLP & 98.4 $\pm$ 1.0 & 0.12 \\
\midrule
Mag-Exact-HR (oracle) & 98.2 $\pm$ 0.9 & 0.06 \\
Mag-Exact-MLP (oracle) & 99.3 $\pm$ 0.2 & 0.06 \\
Mag-Exact-Free (unconstrained oracle) & 34.3 $\pm$ 3.3 & 0.06 \\
\bottomrule
\end{tabular}
}
\end{table}

\paragraph{Main comparison (Table~\ref{tab:main}).} Random probes, the
standard undirected baselines (LapPE on the symmetrized Laplacian, RWSE on
symmetrized random walks), and all $q{=}0$ variants sit at chance,
validating the dataset design: the class signal lives entirely in edge
directions. Sym-RFP obtains $32.4\%$, while Magnetic-RFP reaches
$38.4\%$: access to directed powers alone is not sufficient to match the
learned response families. A fixed magnetic eigenvector PE (Mag-PE with the same potentials
$q$ and a deterministic gauge fix, in the spirit of
\citealp{geisler2023transformers,huang2025good}) does see direction but
reaches only $51.1\%$: without a learnable spectral response, the raw
eigenvector coordinates are a poor interface for the downstream MLP. All
learnable magnetic Krylov families recover the directional structure
($83$--$99\%$). The unconstrained spectral oracle collapses to chance despite
having the most capacity, consistent with the overfitting regime described
by Proposition~\ref{prop:gen}.

\begin{figure*}[t]
\centering
\includegraphics[width=0.24\textwidth]{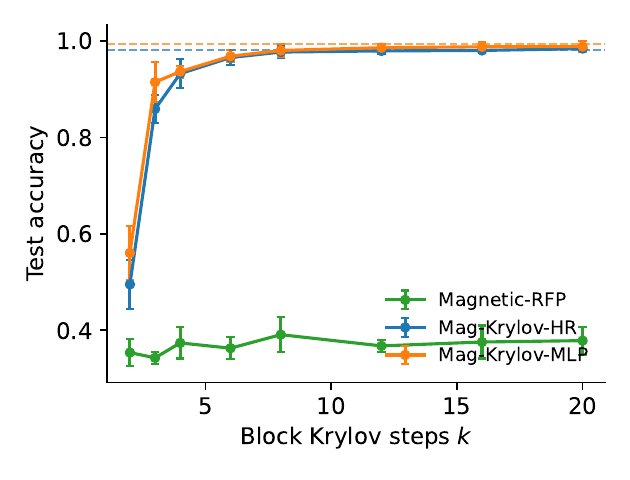}\hfill
\includegraphics[width=0.24\textwidth]{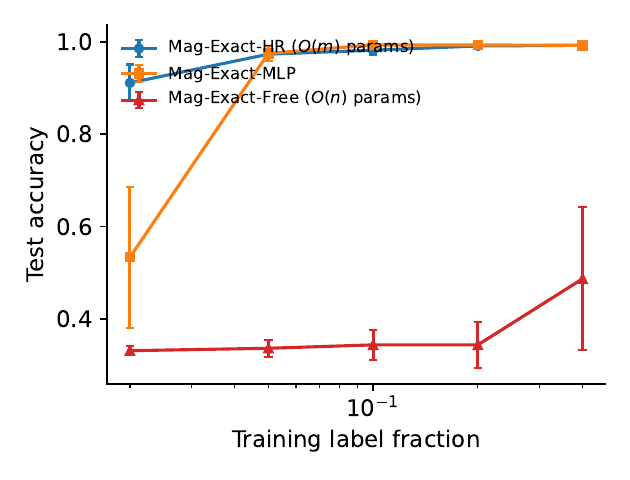}\hfill
\includegraphics[width=0.24\textwidth]{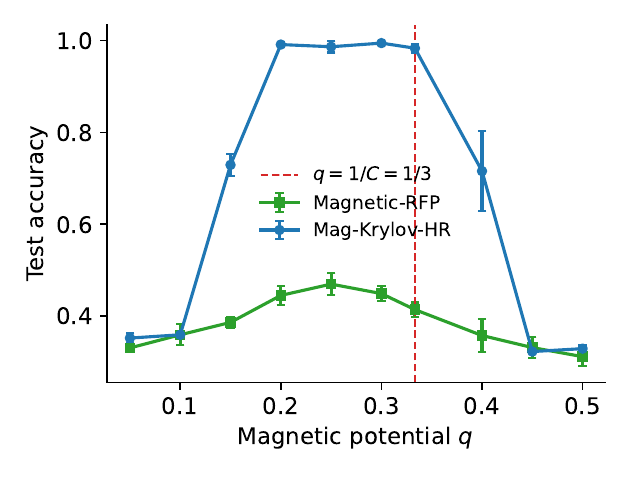}\hfill
\includegraphics[width=0.24\textwidth]{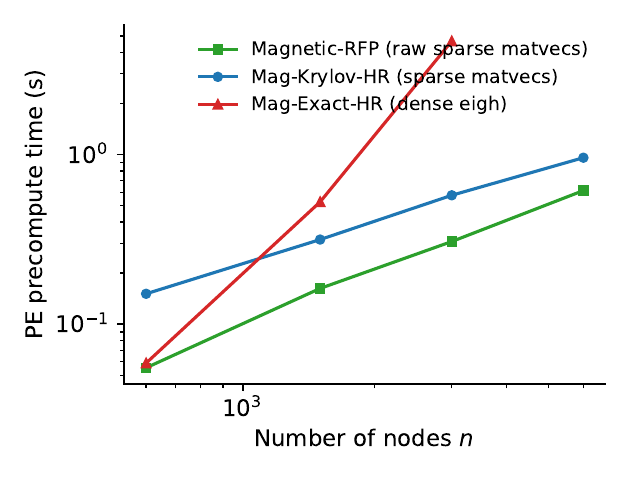}
\caption{DSBM diagnostics, left to right: \emph{(a)}~test accuracy vs.\
block Krylov depth $k$ (dashed lines: exact eigendecomposition oracles), the
experimental counterpart of Theorem~\ref{thm:approx}, together with the
matched Magnetic-RFP power-iterate baseline; \emph{(b)}~label
efficiency in exact mode; the unconstrained per-eigenvalue oracle needs far more labels
than the $O(m)$-parameter heat--resolvent family, consistent with
Proposition~\ref{prop:gen}; \emph{(c)}~single-potential accuracy vs.\ $q$
for learned Krylov and Magnetic-RFP, both responding to the cycle-aligned band
\citep{huang2025good}; \emph{(d)}~PE precompute time vs.\ $n$ at constant
expected degree, measured on the same machine as
Tables~\ref{tab:large_directed_scaling}--\ref{tab:large50k_ablation}; both
Krylov and Magnetic-RFP use sparse matvecs, while dense Hermitian
eigendecomposition is cubic and is skipped beyond $n = 3000$.}
\label{fig:dsbm}
\end{figure*}

\paragraph{Depth sweep (Figure~\ref{fig:dsbm}a).} Accuracy increases
monotonically in $k$: the heat--resolvent variant moves from $49.5\%$ at
$k{=}2$ to $85.9\%$ at $k{=}3$, $97.7\%$ at $k{=}8$, and $98.4\%$ at
$k{=}20$, against an exact oracle at $98.2\%$. Most of the gap is closed by
$k \approx 8$, i.e.\ a basis of $r \le ks \ll n$ dimensions, consistent
with the $O(\log 1/\varepsilon)$ rate of Theorem~\ref{thm:approx}.
Magnetic-RFP stays between $34.3\%$ and $39.1\%$ throughout the same sweep.
Increasing the raw propagation depth alone therefore does not account for
the Krylov-response improvement.

\paragraph{Label efficiency (Figure~\ref{fig:dsbm}b).} With approximation
error removed (exact mode), capacity ordering determines label efficiency in
the direction suggested by Proposition~\ref{prop:gen}: the $O(m)$-parameter
heat--resolvent family reaches $91.1\%$ with only $2\%$ labeled nodes, the
MLP response needs ${\sim}5\%$, and the unconstrained per-eigenvalue oracle stays near
chance even with $20\%$ labels ($34.3\%$) and reach only $48.7\%$ at $40\%$
labels. A separate probe-count sweep (Table~\ref{tab:probe_stability})
confirms the Monte-Carlo side of Proposition~\ref{prop:probe}: increasing
the number of probes from $s{=}8$ to $s{=}32$ roughly halves the standard
deviation and improves accuracy, consistent with the random spectral-kernel
rate established by GIST \citep{rigotti2026gist}.

\paragraph{Potential alignment (Figure~\ref{fig:dsbm}c).} With a single
potential the response is sharply resonant: accuracy is at chance for
$q \le 0.1$, rises through $72.9\%$ at $q{=}0.15$ to a $98$--$99\%$ plateau
for $q \in [0.2, 1/3]$ (the band aligned with the $C{=}3$ cyclic structure),
and collapses back to chance as $q \to 1/2$. Since the informative band is
task-dependent and unknown a priori, a small grid of potentials, each with
its own cheap Krylov cache, inherits the best scale, supporting the
multi-$q$ design of \citet{huang2025good} within a learnable-response
framework.
Magnetic-RFP shows the same directional sensitivity but is substantially
weaker: it rises from $33.1\%$ at $q{=}0.05$ to $47.0\%$ at $q{=}0.25$
before returning to chance near $q{=}1/2$. Thus the magnetic operator exposes
the signal, while the learned spectral response is needed to extract it
reliably.

\paragraph{Scaling (Figure~\ref{fig:dsbm}d).} At constant expected degree
the Krylov precompute is essentially flat ($0.09$s at $n{=}600$ to $0.17$s
at $n{=}50\mathrm{k}$, all three potentials included), while dense Hermitian
eigendecomposition grows cubically ($0.05$s at $n{=}600$, $0.22$s at
$n{=}3000$) and is infeasible at the $50$k scale, where the dense
eigenvectors of a single potential alone would occupy tens of gigabytes.
This is the regime in which the small-graph SPE comparison no longer
applies: exact magnetic spectra become the bottleneck, while Krylov still
uses sparse matvecs and stays below $0.2$ seconds in our diagnostic.
Accuracy at fixed $k{=}10$ drifts only mildly with $n$ ($98.1\%$ at
$n{=}600$ vs.\ $95.6\%$ at $n{=}6000$), recoverable by scaling $k$ (or, at
larger scales, $s$) with the spectral resolution required. Table~\ref{tab:large_directed_scaling} extends this
diagnostic to $50$k-node directed graphs: exact magnetic PEs are skipped as
infeasible, while Mag-Krylov remains well above random and direction-blind
baselines. A 50k-node ablation (Table~\ref{tab:large50k_ablation}) further
separates the two approximation parameters: increasing $k$ from $10$ to $20$ at
fixed $s$ does not improve accuracy, whereas increasing $s$ from $32$ to
$64$ recovers most of the large-graph drop, raising accuracy from $60.7\%$
to $69.8$--$70.6\%$ while keeping preprocessing below one second. This
suggests that, in this regime, the dominant residual error is random-probe
resolution rather than Krylov depth.
Magnetic-RFP also scales through sparse matvecs (from $0.06$s at $n{=}600$
to $0.61$s at $n{=}6000$), but its accuracy remains far below the learned
Krylov response at matched graph sizes.

\begin{table}[t]
\centering
\caption{Real directed graphs (test accuracy, 5 seeds, PE + raw features).}
\label{tab:real}
\resizebox{\columnwidth}{!}{\begin{tabular}{lcccc}
\toprule
PE variant & Chameleon & Cornell & Texas & Wisconsin \\
\midrule
Random probes & \textbf{41.5 $\pm$ 1.9} & 61.3 $\pm$ 2.4 & 63.3 $\pm$ 1.8 & \textbf{73.6 $\pm$ 1.3} \\
Sym-Krylov-HR ($q{=}0$) & 38.1 $\pm$ 1.6 & 61.3 $\pm$ 1.7 & 61.9 $\pm$ 2.4 & 71.8 $\pm$ 0.9 \\
Sym-Krylov-MLP ($q{=}0$) & 30.2 $\pm$ 1.4 & \textbf{63.6 $\pm$ 3.1} & \textbf{65.0 $\pm$ 1.6} & 71.5 $\pm$ 1.5 \\
Mag-Krylov-HR & 30.8 $\pm$ 1.8 & 61.1 $\pm$ 5.4 & 60.8 $\pm$ 2.1 & 72.3 $\pm$ 4.0 \\
Mag-Krylov-MLP & 30.9 $\pm$ 3.9 & 57.2 $\pm$ 3.7 & 61.6 $\pm$ 3.9 & 70.3 $\pm$ 1.7 \\
\bottomrule
\end{tabular}
}
\end{table}

\paragraph{Real directed graphs (Table~\ref{tab:real}).} On WebKB and
Wikipedia networks (loaded \emph{without} symmetrization), all spectral PEs,
directional or not, fail to improve over the random-probe baseline.
Several substantially underperform it on Chameleon. We report this as a
negative result: these benchmarks may not strongly reward the type of global
directional information captured by magnetic spectral PEs under our
experimental setup. This mirrors why
\citet{huang2025good} evaluate on direction-critical tasks (distance
prediction, sorting networks, circuits) rather than node classification. We
therefore complement this node-level diagnostic with the directed pairwise
regression suite in Section~\ref{sec:directed_pairwise}.

\subsection{The Undirected Special Case ($q = 0$)}
\label{sec:undirected}

At $q = 0$ the framework reduces to the real symmetric operator
$A = L - I$, so every framework-level claim can be validated independently
of direction. We summarize experiments run with the undirected instantiation
(synthetic heterophilous graphs, MLP backbone, setup in
Appendix~\ref{app:undirected}).

\begin{table}[t]
\centering
\caption{Undirected controlled comparison (synthetic heterophilous graphs,
test accuracy \%, 10 seeds). Krylov-Cheb reproduces direct Chebyshev
to all reported digits, confirming
Lemma~\ref{lem:exact}; the structured heat family outperforms polynomials
under the same Krylov budget. $\Delta$ is relative to Direct-Cheb.}
\label{tab:undirected_main}
\begin{tabular}{lcc}
\toprule
Method & Acc.\ (\%) & $\Delta$ \\
\midrule
Random probes & 32.83 $\pm$ 1.72 & $-3.24$ \\
Direct-Cheb & 36.07 $\pm$ 4.18 & -- \\
Krylov-Cheb & 36.07 $\pm$ 4.18 & $+0.00$ \\
Krylov-MLP & 33.48 $\pm$ 1.49 & $-2.60$ \\
Krylov-HR & 47.25 $\pm$ 6.87 & $+11.18$ \\
Exact-HR (oracle) & 65.60 $\pm$ 9.53 & $+29.52$ \\
\textbf{Krylov-Heat} & \textbf{72.00 $\pm$ 6.01} & $+35.93$ \\
Exact-Heat (oracle) & 74.77 $\pm$ 7.29 & $+38.70$ \\
\bottomrule
\end{tabular}
\end{table}

\begin{table}[t]
\centering
\caption{Undirected Chebyshev-order sweep with $k = M + 4$ Krylov steps.
Krylov-Cheb matches Direct-Cheb at every order (Lemma~\ref{lem:exact});
Krylov-Heat improves over the best polynomial at the same budget.}
\label{tab:undirected_order}
\resizebox{\columnwidth}{!}{%
\begin{tabular}{ccccc}
\toprule
$M$ & $k$ & Direct-Cheb & Krylov-Cheb & Krylov-Heat \\
\midrule
2  & 6  & 33.62 $\pm$ 1.70 & 33.65 $\pm$ 1.71 & \textbf{68.01 $\pm$ 7.26} \\
4  & 8  & 36.48 $\pm$ 3.75 & 36.48 $\pm$ 3.75 & \textbf{69.85 $\pm$ 6.57} \\
8  & 12 & 36.07 $\pm$ 4.18 & 36.07 $\pm$ 4.18 & \textbf{72.00 $\pm$ 6.01} \\
16 & 20 & 34.61 $\pm$ 3.61 & 34.61 $\pm$ 3.61 & \textbf{73.51 $\pm$ 6.22} \\
\bottomrule
\end{tabular}}
\end{table}

\begin{table}[t]
\centering
\caption{Undirected heterophilous node-classification benchmarks
\textnormal{(ROC-AUC \%, original features, averaged over splits/seeds)}.
The structured Krylov PE improves over no PE and over direct Chebyshev;
LapPE is the strongest fixed baseline on these feature-rich undirected
graphs. RFP here uses the predefined propagation-operator variant of
\citet{eliasof2023graph}. ``--'': the standard sparse-power RWSE implementation densifies
$P^k$ and exceeds memory on Questions ($n{\approx}49$k).}
\label{tab:undirected_real}
\resizebox{\columnwidth}{!}{%
\begin{tabular}{lccc}
\toprule
Method & Minesweeper & Tolokers & Questions \\
\midrule
No PE / MLP & 50.26 $\pm$ 0.98 & 73.49 $\pm$ 0.84 & 66.89 $\pm$ 1.42 \\
LapPE & \textbf{52.46 $\pm$ 1.02} & \textbf{79.92 $\pm$ 0.99} & \textbf{71.48 $\pm$ 1.19} \\
RWSE & 52.42 $\pm$ 1.11 & 77.43 $\pm$ 0.60 & -- \\
RFP & 50.39 $\pm$ 0.16 & 76.47 $\pm$ 0.49 & 67.88 $\pm$ 0.81 \\
Direct-Cheb & 50.29 $\pm$ 1.65 & 72.52 $\pm$ 0.86 & 68.07 $\pm$ 1.13 \\
Krylov-Heat & 51.19 $\pm$ 1.48 & 76.04 $\pm$ 1.05
  & 69.24 $\pm$ 1.61 \\
\bottomrule
\end{tabular}}
\end{table}

Three observations transfer directly. \emph{(i) Exactness in practice:}
Krylov-Cheb and Direct-Cheb agree to all reported digits at every
order (Tables~\ref{tab:undirected_main} and~\ref{tab:undirected_order}),
so the Krylov compression loses nothing on the polynomial family while
providing an orthogonalized, numerically conditioned cache and a projected
operator that can be reused as response parameters change. Both approaches
retain $O(kns)$ large state in the no-deflation case; Krylov additionally
supports non-polynomial matrix functions without rerunning sparse
propagation. \emph{(ii) Structured families under a fixed budget:} the heat family
gains $+36$ points over the best polynomial at identical Krylov cost and
sits within $3$ points of its exact oracle. \emph{(iii) Real data:}
on heterophilous benchmarks (Table~\ref{tab:undirected_real}) Krylov-Heat
improves over no-PE and polynomial baselines on all three datasets, while
LapPE is the strongest method overall. We read this as a property of the
benchmarks rather than of the encodings: these undirected graphs have rich
node features, and a fixed low-pass eigenvector summary suffices; there is
no directional signal for a learnable response to exploit (cf.\ the chance
performance of all symmetric PEs in Table~\ref{tab:main}). The Krylov PE
remains an order of magnitude cheaper to precompute (Questions: $1.2$s
vs.\ $14.5$s for LapPE's sparse \texttt{eigsh}) and carries no
sign/basis-ambiguity caveats. RWSE's sparse-power computation does not
complete on the largest graph. Undirected scaling experiments (up to $n{=}9000$:
Krylov $0.19$s vs.\ exact $30.9$s precompute, with a $4$-point accuracy gap
at fixed $k$) appear in Appendix~\ref{app:undirected}.

\subsection{Directed Pairwise Regression}
\label{sec:directed_pairwise}

\begin{table*}[htbp]
\centering
\caption{Resource trade-off between spectral pairwise readouts and our
Krylov-probe readout. $P$ is the number of queried node pairs, $s$ the number
of probes, $H$ the number of response heads, and $k_{\mathrm{eig}}$ the number
of retained eigenvectors. Partial-eigensolver cost depends on iterations and
spectral gaps and, like Krylov, uses sparse matvecs. Measured times are from the directed
large-graph diagnostic
(Tables~\ref{tab:large_directed_scaling}--\ref{tab:large50k_ablation}).}
\label{tab:resource_tradeoff}
\resizebox{\textwidth}{!}{%
\begin{tabular}{lccc}
\toprule
Method & Precompute & Stored PE state & Pairwise readout \\
\midrule
Full-spectrum oracle
  & $O(Q n^3)$ dense eigensolve
  & $O(Qn^2)$
  & $O(P Q n^2)$ \\
Truncated Mag-PE + SPE
  & sparse partial eigensolver
  & $O(Q n k_{\mathrm{eig}})$
  & $O(P Q k_{\mathrm{eig}}^2)$ \\
Mag-Krylov, learnable cache
  & $O(Qk(\mathrm{nnz}\,s + nks^2))$
  & $O(Qnks + Qk^2s^2)$
  & $O(P Q H^2 s)$ \\
Mag-Krylov, frozen response
  & no new graph preprocessing
  & $O(QHns)$ or $O(nd_{\rm PE})$
  & $O(P Q H^2 s)$ \\
\midrule
Measured scaling
  & full oracle skipped at $n{>}3000$; Krylov $0.17$--$0.76$s at $50$k
  & full basis vs.\ Krylov blocks
  & same cached probes \\
\bottomrule
\end{tabular}
}
\end{table*}

We evaluate on the directed shortest-path-distance (SPD) benchmark of
\citet{huang2025good}: random connected DAGs ($16$--$63$ nodes for
training, $64$--$71$ for testing), where the model regresses
$\mathrm{spd}(i,j)$ for all reachable pairs. We integrate our encoder into
their public framework and keep dataset, predictor head, batch size, and
epochs identical across methods ($40$k training graphs; $3$ seeds). Their
headline method couples multi-$q$ magnetic eigenvectors with the SPE
invariant network \citep{huang2024stability}. It retains the lowest
$k_{\rm eig}$ eigenpairs and forms per-pair products of their entries.
The released preprocessing code uses a dense decomposition on these small
graphs, but the method itself admits a sparse partial eigensolver; its cost
therefore depends on $k_{\rm eig}$, matvec iterations, and spectral gaps.

Following the random spectral-kernel construction of GIST
\citep{rigotti2026gist}, for any pair $(i,j)$ we form the normalized probe
inner-product estimator
$\frac{1}{s\sigma^2}\sum_{t=1}^{s}(h_a(\mathbf{L}_q)\mathbf{z}_t)_i
\overline{(h_b(\mathbf{L}_q)\mathbf{z}_t)_j}$. It is an unbiased Hutchinson
estimate of the eigenbasis-independent matrix entry
$(h_a h_b^{*})(\mathbf{L}_q)_{ij}$, computed from the \emph{same}
precomputed probes as the node-level PE at no extra preprocessing cost.
Relative to GIST's same-response real symmetric kernel, we use cross-response
products on complex magnetic operators and multiple potentials.
Table~\ref{tab:spd} isolates the contribution of each design axis: the
pairwise readout (vs.\ node-level features alone), the response-family
capacity $H$, the cross-head products ($H^2$ effective responses), and the
probe count $s$ controlling the $O(1/\sqrt{s})$ Monte-Carlo error of
Proposition~\ref{prop:probe}.

\begin{table}[t]
\centering
\caption{Directed SPD regression (test MSE, mean$\pm$std over $3$ seeds)
on the framework of \citet{huang2025good}. All rows share the magnetic
operator with $Q{=}10$ potentials; ours additionally shares one set of $s$
probes across node-level and pairwise features.}
\label{tab:spd}
\resizebox{\columnwidth}{!}{%
\begin{tabular}{lc}
\toprule
Configuration & Test MSE \\
\midrule
Mag-PE eigvecs + MLP encoder & $0.314 \pm 0.003$ \\
Mag-Krylov, node-level only ($H{=}4$, $s{=}8$) & $0.272 \pm 0.008$ \\
\;+ pairwise readout & $0.127 \pm 0.003$ \\
\;+ capacity $H{=}16$ & $0.069 \pm 0.002$ \\
\;+ probes $s{=}32$ & $0.051 \pm 0.001$ \\
\;+ cross-head products ($H^2$) & $0.042 \pm 0.001$ \\
\midrule
Mag-PE + SPE ($k_{\rm eig}{=}32$) & $0.0037 \pm 0.0004$ \\
\bottomrule
\end{tabular}}
\end{table}

Each row corresponds to a quantity in our analysis, and each
improves the error in the predicted direction. The probe-based pairwise
readout halves the error twice
over; enlarging the response family helps until the Monte-Carlo noise
floor binds; and quadrupling $s$ buys the $\approx 2\times$ noise
reduction the $1/\sqrt{s}$ rate predicts, with diminishing returns
thereafter (at $s{=}128$ the error improves only to $\approx 0.033$ at
$4\times$ the memory). The truncated SPE baseline remains an order of
magnitude better on these small graphs: SPD demands highly accurate pairwise
resolvent information, and $k_{\rm eig}{=}32$ already captures much of it.
The comparison motivates the controlled $1/\sqrt{s}$ approximation path,
but does not by itself establish a sparse-matvec advantage because partial
eigensolvers also use sparse matvecs.

We next keep the same protocol and evaluate the full directed pairwise
suite: shortest-path distance (SPD), longest-path distance (LPD), and the
step-4 walk-profile vector (WP). These targets stress different global
properties of the directed DAGs, while reusing the same encoder
implementation and the same pairwise cross-head readout. Table~\ref{tab:directed_pairwise}
shows that the Krylov matrix-function PE remains far stronger than the
standard Mag-PE encoder, improving the error by $7.5\times$ on SPD,
$2.1\times$ on LPD, and $7.4\times$ on WP. SPE remains the best method on
average, but its variance is much larger on the harder LPD/WP targets: in
both cases one seed is close to, or worse than, the structured Krylov model.

\begin{table}[t]
\centering
\caption{Directed pairwise regression suite (test MSE, mean$\pm$std over
$3$ seeds). All tasks use random connected DAGs from
\citet{huang2025good}; lower is better.}
\label{tab:directed_pairwise}
\resizebox{\columnwidth}{!}{%
\begin{tabular}{lccc}
\toprule
Method & SPD & LPD & WP \\
\midrule
Mag-PE eigvecs + MLP encoder
  & $0.314 \pm 0.003$ & $0.446 \pm 0.004$ & $2.978 \pm 0.006$ \\
Mag-Krylov + pairwise cross
  & $0.042 \pm 0.001$ & $0.210 \pm 0.004$ & $0.401 \pm 0.005$ \\
Mag-PE + SPE ($k_{\rm eig}{=}32$)
  & $0.0037 \pm 0.0004$ & $0.102 \pm 0.058$ & $0.306 \pm 0.132$ \\
\bottomrule
\end{tabular}}
\end{table}

Thus we do not interpret Tables~\ref{tab:spd}--\ref{tab:directed_pairwise}
as a small-graph leaderboard against SPE. Truncated eigenvector methods are
the right tool when the graph is small enough and the task demands highly
accurate pairwise spectral entries. The contribution is instead a controlled approximation
path: the same eigenbasis-independent matrix-function PE supports node features and
pairwise readouts, exposes a probe count $s$ with $1/\sqrt{s}$ error, and
trades some accuracy for full-spectrum learnable magnetic filtering without
extracting or stabilizing individual eigenvectors.

\section{Discussion and Limitations}

Our diagnostic benchmark is intentionally adversarial to symmetrization;
real graphs mix directional and undirected signal, where the margin between
$q{=}0$ and magnetic PEs naturally narrows. The potentials $\vec q$ are
currently a fixed grid: a learnable $q$ requires differentiating through the
operator and rebuilding the Krylov cache, an interesting direction we leave
open. Our theory covers approximation and a covering-number capacity bound, not
the optimization dynamics of the response families; the RMS normalization we
found necessary in practice (Section~\ref{sec:exp}) suggests conditioning of
the spectral parameterization deserves study in its own right. Finally,
extending beyond the pairwise regression benchmarks of
Section~\ref{sec:directed_pairwise} to the circuit and sorting-network tasks
of \citet{huang2025good}, which require transformer backbones, is the
natural next experimental step.

\section{Conclusion}

Learnable spectral PEs for directed graphs do not require eigenvectors. A
matrix-function formulation over magnetic operators is independent of
eigenbasis choices, admits uniform Krylov approximation guarantees with
$O(\log 1/\varepsilon)$ sparse steps, and separates the design into three
independent parameters: the Krylov depth $k$ (approximation bias), the
response-family capacity, and the probe count $s$ (Monte-Carlo resolution).
This decomposition, validated here on directed graphs, is not specific to
the directed setting.

\FloatBarrier

\bibliographystyle{plainnat}
\bibliography{references}

\appendix

\section{Proofs}
\label{app:proofs}

\subsection{Proposition~\ref{prop:gauge}}
A matrix function of a Hermitian matrix is defined by the spectral calculus
$h(A) = \sum_i h(\lambda_i) \Pi_i$ where $\Pi_i$ are the (unique) orthogonal
spectral projectors; any eigenbasis choice within an eigenspace yields the
same $\Pi_i$, hence the same $h(A)$ and the same $Z = h(A)R$. Permutation
equivariance follows from $h(PAP^\top) = P h(A) P^\top$ for unitary $P$. The
Krylov statement follows because $Q$ is constructed from
$\{A^j R\}_{j<k}$: replacing $(A, R) \mapsto (PAP^\top, PR)$ maps
$Q \mapsto PQ$ (up to the same internal QR conventions), leaving
$T = Q^{\mathsf H} A Q$ and $G = Q^{\mathsf H} R$ unchanged and mapping
$\widehat Z \mapsto P \widehat Z$. \qed

\subsection{Proposition~\ref{prop:probe}}
For the conditional equivariance claim, use the identity
$h(PAP^\top)=P h(A)P^\top$ and multiply by $PR$. For distributional
equivariance, an iid complex Gaussian matrix satisfies $P^\top R'
\stackrel{d}{=} R$, hence
$h(PAP^\top)R' = P h(A) P^\top R' \stackrel{d}{=} P h(A)R$.

For the pairwise estimator, write the $i$th row of $F_a$ as
$u^{\mathsf H}=e_i^{\mathsf T}F_a$ and the $j$th row of $F_b$ as
$v^{\mathsf H}=e_j^{\mathsf T}F_b$. A single probe contributes
$
X_t=(u^{\mathsf H}r_t)\overline{(v^{\mathsf H}r_t)}.
$
Since $\E r_t r_t^{\mathsf H}=\sigma^2 I$,
\[
\E X_t
= u^{\mathsf H}(\E r_t r_t^{\mathsf H})v
= \sigma^2 u^{\mathsf H}v
= \sigma^2 (F_aF_b^{\mathsf H})_{ij}.
\]
Thus $(s\sigma^2)^{-1}\sum_t X_t$ is unbiased. For the variance, Isserlis'
formula for circular complex Gaussians gives
\begin{align*}
\E |X_t|^2
&= \E |u^{\mathsf H}r_t|^2 |v^{\mathsf H}r_t|^2 \\
&= \sigma^4\big(\norm{u}_2^2\norm{v}_2^2
   + |u^{\mathsf H}v|^2\big) \\
&\le 2\sigma^4\norm{u}_2^2\norm{v}_2^2 .
\end{align*}
After subtracting $|\E X_t|^2$ and averaging independent probes,
\[
\mathrm{Var}(\widehat K_{ab}(i,j))
\le \frac{2}{s}\,
\norm{e_i^{\mathsf T}F_a}_2^2\norm{e_j^{\mathsf T}F_b}_2^2 .
\]
\qed

\subsection{Lemma~\ref{lem:exact}}
By induction: $A^j R \in \mathcal{K}_k$ for $j \le k-1$, and on
$\mathcal{K}_k$ the compression $T = Q^{\mathsf H} A Q$ satisfies
$Q T^j Q^{\mathsf H} R = A^j R$ for $j \le k-1$ (standard block Lanczos
argument; rank deflation only shrinks the space when it is already
invariant). Linearity extends to polynomials. \qed

\subsection{Proposition~\ref{prop:walk}}
Choose $a=\min(m,k-1)$ and $b=m-a$; for every $m\le 2k-2$ both exponents
lie in $\{0,\ldots,k-1\}$. Lemma~\ref{lem:exact} reproduces
$A_q^aR$ and $A_q^bR$ exactly. Since $A_q$ is Hermitian,
$A_q^a(A_q^b)^{\mathsf H}=A_q^{a+b}=A_q^m=(-1)^mB_q^m$.
Proposition~\ref{prop:probe} therefore gives unbiasedness and
$O(s^{-1/2})$ standard deviation entrywise. Multiplication by $(-1)^m$
therefore produces an unbiased estimator of $(B_q^m)_{ij}$. Finally,
\citet{huang2026powers} show that the values $(B_q^m)_{ij}$ on the stated
frequency grid are the discrete Fourier transform of the degree-normalized
length-$m$ directed walk profile. Applying the inverse transform proves the
claim; as a fixed finite-dimensional linear map, it preserves the
$O(s^{-1/2})$ rate up to an $m$-dependent constant. \qed

\subsection{Theorem~\ref{thm:approx}}
Fix $\theta$ and any polynomial $p$ with $\deg p \le k-1$. By
Lemma~\ref{lem:exact},
\begin{align*}
&\norm{h_\theta(A)R - Q h_\theta(T) G}_F \\
&\quad\le \norm{(h_\theta - p)(A) R}_F + \norm{Q (h_\theta - p)(T) G}_F \\
&\quad\le \big( \norm{(h_\theta - p)(A)}_2 + \norm{(h_\theta - p)(T)}_2 \big)
\norm{R}_F \\
&\quad\le 2 \sup_{x \in [-1,1]} |h_\theta(x) - p(x)| \cdot \norm{R}_F,
\end{align*}
using $\spec(T) \subseteq [\lambda_{\min}(A), \lambda_{\max}(A)] \subseteq
[-1,1]$ (Cauchy interlacing for compressions) and the spectral mapping
theorem. Taking the infimum over $p$ and the supremum over $\theta$ gives the
first claim. For the rates: each resolvent $x \mapsto (x + 1 + \tau)^{-1}$ is
analytic on a Bernstein ellipse with parameter
$\rho^{-1} = 1 + \tau + \sqrt{\tau^2 + 2\tau}$, giving
$E_{k-1} \le C_\tau \rho^k$ \citep[Ch.~8]{trefethen2019approximation};
each heat term $x \mapsto e^{-t(x+1)}$ is entire, with
$E_{k-1}(e^{-t(\cdot+1)}) \le C e^{-k^2/(2t)}$-type super-geometric decay for
$k \gtrsim t$. Sub-additivity of $E_{k-1}$ over the mixture with $\ell_1$
coefficient bound $B$ yields the uniform bound: the heat parameters lie in
the compact interval $[t_{\min},t_{\max}]$, while increasing $\tau$ moves the
resolvent pole farther from $[-1,1]$, so the worst resolvent approximation
rate over $\tau\ge\tau_{\min}$ occurs at $\tau_{\min}$.
\qed

\subsection{Proposition~\ref{prop:gen}}
Let $\Theta \subset \mathbb{R}^{d}$ be a compact parameter set for a response
family, and suppose $\theta\mapsto h_\theta(\xi)$ is $L_h$-Lipschitz
uniformly over $\xi\in[-1,1]$. Because the Krylov cache is fixed during
training,
\[
\widehat Z(\theta)-\widehat Z(\theta')
=Q\big(h_\theta(T)-h_{\theta'}(T)\big)G
\]
is Lipschitz in $\theta$ with constant at most
$L_h\norm{Q}_2\norm{G}_F$. Composing with any member of the fixed downstream
class and a bounded uniformly Lipschitz loss preserves Lipschitzness up to
constants. Therefore an
$\varepsilon$-net of $\Theta$ induces a $C\varepsilon$-net of the loss class.
For a bounded $d$-dimensional compact parameter set, the covering number is
\[
\mathcal{N}(\varepsilon,\Theta,\norm{\cdot}_2)\le (C/\varepsilon)^d,
\qquad
\log\mathcal{N}=O(d\log(C/\varepsilon)).
\]
Taking the product with an $\varepsilon$-net of the downstream class adds
the common term $\mathcal C_{\rm down}(\varepsilon)$ to both response
families.
Standard covering-number bounds for bounded losses then yield uniform
deviation
\[
\sup_{f\in\mathcal{F}} |R(f)-\widehat R_\ell(f)|
=\widetilde O\!\left(\sqrt{d/\ell}\right).
\]
For heat--resolvent mixtures, $d=O(Qm)$ (a constant number of parameters per
component and per potential). For an unconstrained per-eigenvalue oracle over
a rank-$r$ spectral representation, $d=Qr$, with $r$ as large as $n$.
Applying the same upper covering bound gives the two rates in the proposition
and explains the observed
low-label overfitting of the oracle; it is not a statement about LLPE's
shared Chebyshev parameterization. \qed

\section{Undirected experimental details}
\label{app:undirected}

The undirected experiments of Section~\ref{sec:undirected} use the real
symmetric operator $A = L - I = -D^{-1/2} \mathrm{Adj}\, D^{-1/2}$ (the
$q{=}0$ instantiation) and random Gaussian probes shared across all
variants. Synthetic graphs are heterophilous SBMs with $n = 1200$,
$3$ classes, $p_{\mathrm{in}} = 0.008$, $p_{\mathrm{out}} = 0.035$
(probabilities scaled $O(1/n)$ in the scaling study), \emph{no} node
features (PE-only classification); $s = 32$ probes, PE dimension $64$,
$8$ heads, $8$ heat/resolvent components, $2$-layer MLP backbone with
hidden width $128$, dropout $0.1$, AdamW with learning rate $10^{-3}$ and
weight decay $10^{-5}$, up to $500$ epochs with patience $150$ and
best-validation restoration. Controlled-comparison results average 10
seeds, scaling results 3 seeds; heterophilous benchmarks use the standard
splits of the respective datasets with original features and ROC-AUC for
the binary tasks. All synthetic numbers were re-verified by rerunning the
released configuration end to end.
``Krylov-Heat'' denotes the heat-only structured family;
``Krylov-HR'' the heat--resolvent mixture; oracles use a full
eigendecomposition. The Chebyshev-order sweep ties the Krylov budget to the
polynomial order via $k = M + 4$ so that polynomial exactness
(Lemma~\ref{lem:exact}) is guaranteed with margin.

\begin{table}[t]
\centering
\caption{Undirected scaling with graph size (test accuracy \%, 3 seeds;
edge probabilities scaled $O(1/n)$ for constant average degree). Exact uses
a full eigendecomposition.}
\label{tab:undirected_scaling}
\resizebox{\columnwidth}{!}{%
\begin{tabular}{llccc}
\toprule
$n$ & Method & Acc.\ (\%) & Precompute & Train time \\
\midrule
1200 & Direct-Cheb       & 36.11 $\pm$ 1.73 & 0.006s & 2.01s \\
1200 & Krylov-Heat & \textbf{69.44 $\pm$ 0.66} & 0.034s & 2.07s \\
1200 & Exact-Heat  & 71.75 $\pm$ 1.25 & 0.043s & 4.14s \\
\midrule
3000 & Direct-Cheb       & 54.89 $\pm$ 8.71 & 0.014s & 6.29s \\
3000 & Krylov-Heat & \textbf{81.08 $\pm$ 1.46} & 0.077s & 5.18s \\
3000 & Exact-Heat  & 81.14 $\pm$ 0.59 & 0.528s & 33.98s \\
\midrule
5000 & Direct-Cheb       & 66.74 $\pm$ 8.13 & 0.019s & 8.36s \\
5000 & Krylov-Heat & \textbf{86.11 $\pm$ 1.78} & 0.102s & 6.67s \\
5000 & Exact-Heat  & 89.39 $\pm$ 0.79 & 3.441s & 96.14s \\
\midrule
9000 & Direct-Cheb       & 73.61 $\pm$ 4.39 & 0.033s & 12.55s \\
9000 & Krylov-Heat & \textbf{87.04 $\pm$ 0.36} & 0.192s & 12.81s \\
9000 & Exact-Heat  & 91.01 $\pm$ 0.54 & 30.926s & 310.16s \\
\bottomrule
\end{tabular}}
\end{table}

\begin{table}[t]
\centering
\caption{Random-probe stability on the directed DSBM at $n{=}3000$ (test
accuracy \%, 10 seeds). Increasing the probe count from $s{=}8$ to $s{=}32$
reduces the empirical standard deviation by roughly the predicted
$1/\sqrt{s}$ factor and improves accuracy.}
\label{tab:probe_stability}
\begin{tabular}{lc}
\toprule
Probe count & Accuracy \\
\midrule
$s=8$  & $73.31 \pm 7.01$ \\
$s=32$ & $86.86 \pm 3.23$ \\
\bottomrule
\end{tabular}
\end{table}

\begin{table}[t]
\centering
\caption{Large directed DSBM scaling (test accuracy \%, 3 seeds). Edge
probabilities are scaled as $O(1/n)$ to keep average degree constant.
Exact magnetic eigendecomposition is skipped for all rows because
$n>\texttt{exact\_max\_n}=3000$.}
\label{tab:large_directed_scaling}
\resizebox{\columnwidth}{!}{%
\begin{tabular}{lccc}
\toprule
$n$ & Random & Sym-Krylov-HR & Mag-Krylov-HR \\
\midrule
10k & $33.56 \pm 0.39$ & $33.60 \pm 0.21$ & $\mathbf{79.02 \pm 2.46}$ \\
20k & $33.64 \pm 0.05$ & $33.61 \pm 0.33$ & $\mathbf{73.88 \pm 4.24}$ \\
50k & $33.24 \pm 0.21$ & $33.26 \pm 0.05$ & $\mathbf{60.72 \pm 1.59}$ \\
\bottomrule
\end{tabular}}
\end{table}

\begin{table}[t]
\centering
\caption{50k-node directed DSBM ablation (test accuracy \%, 3 seeds). Exact
magnetic PE is skipped in every row. Increasing probes improves accuracy far
more than increasing Krylov depth alone. Times are measured on the same
machine as Figure~\ref{fig:dsbm}d.}
\label{tab:large50k_ablation}
\begin{tabular}{lcc}
\toprule
Setting & Mag-Krylov-HR & Precompute \\
\midrule
$k{=}10,\ s{=}32$ & $60.72 \pm 1.59$ & $0.17$s \\
$k{=}20,\ s{=}32$ & $60.52 \pm 1.66$ & $0.38$s \\
$k{=}10,\ s{=}64$ & $69.78 \pm 2.70$ & $0.33$s \\
$k{=}20,\ s{=}64$ & $\mathbf{70.62 \pm 2.73}$ & $0.76$s \\
\bottomrule
\end{tabular}
\end{table}

\section{Implementation details}
\label{app:impl}
Block Lanczos uses QR with rank revealing via $|R_{ii}|$ thresholding
(tolerance $10^{-7}$) and two-pass reorthogonalization; we log
orthogonality error $\norm{Q^{\mathsf H} Q - I}_F / \sqrt{r}$
(${\sim}10^{-7}$ in float32), Hermiticity error of $T$, and the projection
residual $\norm{AQ - QT}_F / \norm{AQ}_F$.

\paragraph{Directed experiments.} Cyclic DSBM with $n = 600$ (except the
scaling study), $C = 3$ classes, $p_f = 0.05$, $p_b = 0.005$, no node
features; potentials $\vec q = (0, \tfrac16, \tfrac13)$, $s = 32$ complex
Gaussian probes, $k = 10$ block steps (except the depth sweep), PE
dimension $32$, $4$ heads, $6$ heat/resolvent components, MLP responses
with $8$ fixed Fourier frequencies; stratified splits with $10\%$ train /
$20\%$ validation. Training: AdamW, lr $2\cdot10^{-3}$, weight decay
$10^{-4}$, dropout $0.5$, up to 300 epochs with patience 50 on validation
accuracy, model restored to the best epoch. Per-head RMS normalization
with learnable gain is applied to all response families
(Section~\ref{sec:method}); disabling it degrades Cheb and MLP families
drastically (e.g.\ Mag-Krylov-Cheb $34\% \to 93\%$ with normalization)
while leaving the internally normalized heat--resolvent family unchanged,
which is why we treat it as part of the parameterization rather than a
tunable trick. The backbone is a 2-layer MLP on $[X; \mathrm{PE}]$; all
variants share probes, splits, seeds, and backbone. Code is in the
supplement. Large-scale directed runs use the same directed SBM with
edge probabilities scaled as $1/n$ to keep expected degree constant:
$(p_f,p_b)=(0.0024,0.00030)$ for $n{=}10$k, $(0.0012,0.00015)$ for
$n{=}20$k, and $(0.00048,0.00006)$ for $n{=}50$k. The large-scale table
uses $80$ epochs with patience $20$ and skips exact variants above
$n=3000$. The probe-stability table fixes $n{=}3000$ and compares
$s{=}8$ vs.\ $s{=}32$ over 10 seeds.

\end{document}